\documentclass[conference]{IEEEtran}

\usepackage{microtype}
\usepackage{graphicx}
\usepackage{subfigure}
\usepackage{booktabs} 
\usepackage{amsmath}
\usepackage{amssymb}
\usepackage{bbm}
\usepackage{bm}
\usepackage{dsfont}
\usepackage[list=off]{caption}
\usepackage{xspace} 
\usepackage{scalefnt}
\usepackage{algpseudocode}
\usepackage{textcomp}
\usepackage{enumitem}
\usepackage[table]{xcolor}
\def\BibTeX{{\rm B\kern-.05em{\sc i\kern-.025em b}\kern-.08em
    T\kern-.1667em\lower.7ex\hbox{E}\kern-.125emX}}

\usepackage{hyperref}

\usepackage{mathtools}
\DeclarePairedDelimiter\abs{\lvert}{\rvert}%
\DeclarePairedDelimiter\norm{\lVert}{\rVert}%

\DeclareMathOperator{\argmaxG}{argmax}

\definecolor{myred}{rgb}{0.8,0,0}
\definecolor{mygreen}{rgb}{0,0.6,0}
\definecolor{myblue}{rgb}{0,0,0.7}
\definecolor{mygray}{rgb}{0.4,0.4,0.4}
\definecolor{mygreen2}{rgb}{0.0,0.3,0.0}

\newcommand{\rl}{{\sc rl}\xspace}
\newcommand{\lp}{{\sc lp}\xspace}
\newcommand{\gcp}{{\sc gcp}\xspace}

\newcommand{\uvfa}{{\sc uvfa}\xspace}

\newcommand{\uvfas}{{\sc uvfa}s\xspace}
\newcommand{\euvfa}{{\sc e-uvfa}\xspace}

\newcommand{\ddqn}{double {\sc dqn}\xspace}
\newcommand{\dqn}{{\sc dqn}\xspace}
\newcommand{\dqnfd}{{\sc dqn}{\tt f}{\sc d}\xspace}
\newcommand{\ddpgfd}{{\sc ddpg}{\tt f}{\sc d}\xspace}
\newcommand{\curious}{{\sc curious}\xspace}
\newcommand{\clic}{{\sc clic}\xspace}
\newcommand{\clicrnd}{{\sc clic-rnd}\xspace}
\newcommand{\eh}{$\mathbf{E_h}$\xspace}
\newcommand{\esix}{$\mathbf{E_6}$\xspace}
\newcommand{\eun}{$\mathbf{E_1}$\xspace}
\newcommand{\etrois}{$\mathbf{E_3}$\xspace}

\begin{document}

\title{CLIC: Curriculum Learning and Imitation for object Control in non-rewarding environments}

\author{\IEEEauthorblockN{Pierre Fournier}
\IEEEauthorblockA{\textit{Institut des Syst\`emes Intelligents et de Robotique} \\
\textit{Sorbonne Universit\'e, CNRS UMR 7222}\\
Paris, France \\
pierre.fournier@isir.upmc.fr}
\and
\IEEEauthorblockN{C\'edric Colas}
\IEEEauthorblockA{\textit{FLOWERS Team} \\
\textit{INRIA Bordeaux Sud Ouest}\\
Talence, France \\
Cedric.Colas@inria.fr}
\and
\IEEEauthorblockN{Mohamed Chetouani}
\IEEEauthorblockA{\textit{Institut des Syst\`emes Intelligents et de Robotique} \\
\textit{Sorbonne Universit\'e, CNRS UMR 7222}\\
Paris, France \\
Mohamed.Chetouani@sorbonne-universite.fr}
\and
\IEEEauthorblockN{Olivier Sigaud}
\IEEEauthorblockA{\textit{Institut des Syst\`emes Intelligents et de Robotique} \\
\textit{Sorbonne Universit\'e, CNRS UMR 7222}\\
Paris, France \\
Olivier.Sigaud@upmc.fr}
}

\maketitle


\begin{abstract}
In this paper we study a new reinforcement learning setting where the environment is non-rewarding, contains several possibly related objects of various controllability, and where an apt agent Bob acts independently, with non-observable intentions. We argue that this setting defines a realistic scenario and we present a generic discrete-state discrete-action model of such environments. To learn in this environment, we propose an unsupervised reinforcement learning agent called CLIC for Curriculum Learning and Imitation for Control. CLIC learns to control individual objects in its environment, and imitates Bob's interactions with these objects. It selects objects to focus on when training and imitating by maximizing its learning progress. We show that CLIC is an effective baseline in our new setting. It can effectively observe Bob to gain control of objects faster, even if Bob is not explicitly teaching. It can also follow Bob when he acts as a mentor and provides ordered demonstrations. Finally, when Bob controls objects that the agent cannot, or in presence of a hierarchy between objects in the environment, we show that CLIC ignores non-reproducible and already mastered interactions with objects, resulting in a greater benefit from imitation. 
\end{abstract}



\section{Introduction}
\label{sec:introduction}

For now, deep reinforcement learning (\rl) is mostly used to perform end-to-end learning of a single task in an environment. It leverages rewards provided by the environment to extract good representations from raw sensorimotor data and learns optimal policies from these representations to solve the task.

This scenario is of limited use for real life reinforcement learning. Instead, an environment may not contain any predefined task, nor extrinsically rewarding signals. In this paper, we propose to rather seek to control the environment through the discovery of purposeful behaviors. More precisely, most environments contain numerous separately evolving objects, that change states when acted upon. Drawing a parallel with how infants explore their surroundings, it seems sensible to explore such environments by attempting to act on and control these objects, individually or together. In this work, a control objective corresponds to being able to set an object individual state to a desired value. 

A naive strategy to learn in such a context would be to randomly choose an object at the start of each episode and explore how to act upon it during the episode. Two issues arise with such a strategy. 

First, if the environment contains many objects, random exploration may learn very slowly and cost many samples. For example, take an object that requires a precise and long sequence of actions to set it in some configuration. When the agent acquires the first successful trajectories for this object, it may be in a learning state prone to reproduce these successes, but the latter will be lost with its next updates if the agent focus shifts away from this object in the next episode. 

Second, environments exhibit structure in the sense that some objects may be more difficult to control than others, not controllable, or interdependent. Take the example of the classic ``shape sorting cube'' developmental toy: manipulating objects of each shape can be an objective at first, before trying to insert one shape in the sorting cube, then another, and so on. Intuitively, picking random objects to practice is likely to be suboptimal: some may not be controllable, or become so only after having mastered others, while training too much on already controlled ones may slow down progress on new ones. 

In other words, random exploration in an environment with many objects possibly in a structure is probably sample-inefficient. Instead, the agent should learn and follow a curriculum, both to focus and leverage the environment structure. 

Another natural strategy to improve sample efficiency in such an environment consists in narrowing the exploration by copying other agents' behaviors. Imitation learning and \rl have been combined and studied under the Learning from Demonstrations (LfD) paradigm \cite{Schaal_ANIPS_1997}. Usually, two hypotheses are made: 1) an expert provides demonstrations to guide the agent, 2) it does so for one task only, that the agent does not have to identify. Once again, in a realistic setting, these assumptions may not hold: the other agent can act independently of the learner and perform behaviors with various intentions, that are not readily observable. While the agent may not be teaching, its behaviors can result in coincidental demonstrations for various ways to control objects: cleaning up the room will likely result in demonstrations for how to manipulate toys, for instance. 

To leverage these fortuitous demonstrations, an \rl agent needs to identify in its own control objectives what the other agent's trajectories can help learn, and reevaluate them accordingly. For all that, the agent may not benefit from trying to reproduce all the behaviors it observes: some of them may touch upon parts of the environment that the agent cannot control (because its actions are limited for example), or the controls shown may be too hard/easy to reproduce given the current level of the agent. Once again, curriculum learning should be used to avoid wasting samples.

In the previous paragraphs, we have described a new challenging reinforcement learning setting where:
\begin{itemize}
    \item the environment does not contain any predefined task and does not provide external rewards, but contains a number of separately available objects, potentially in a structure.
    \item an other unconstrained agent interacts independently with the objects, coincidentally providing demonstrations for the control objectives of the learner.
    \item the agent must explore autonomously and imitate them by following a learned curriculum so as to gain maximum control of surrounding objects and do so faster than with random exploration and imitation.
\end{itemize} 

The challenges for the agent in this setting are three-fold: (i) purposefully controlling parts only of the state space (namely objects), (ii) extracting from an other agent's behaviors the control objectives they help learn (objects for which they contain fortuitous control demonstrations), and (iii) devising a curriculum learning strategy to leverage the structure of the environment and improve sample-efficiency. 

This work first proposes a generic and accessible discrete-state discrete-action model of the type of environment described hitherto, and where the three aforementioned challenges remain. To address them, we design an agent named \clic that extends goal-conditioned policies to tackle finer control of the state space, uses a simple heuristic to identify behaviors from another agent that help achieve its own control objectives, and maximizes absolute learning progress when choosing what to imitate in such behaviors and what to explore next.

The results show that \clic meets the three challenges: it features independent reusable control of the environment objects when possible, and effectively leverages its curriculum to identify then ignore non-reproducible or already mastered behaviors, both when imitating and when exploring. Additionally, when combining imitation and curriculum, it exhibits the interesting property that it can be taught control of objects in a certain order by simply showing it demonstrations for the objects in the same order. This closely matches the way a human caregiver would influence the order in which an infant discovers some toys, for example.

\section{Related work}

This work investigates unexplored questions at the intersection of unsupervised \rl, imitation and curriculum learning.

\subsection{Reinforcement learning without external rewards}

Several \rl agents that do not receive any reward from the environment have been recently proposed, often called "unsupervised" or "self-supervised" \rl agents. Instead the reinforcement signal can come from the discovery of skills with maximal diversity \cite{eysenbach2018diversity}, or options maximizing mutual information with respect to their end states \cite{gregor2016variational,warde2018unsupervised}, or from identifying incremental \emph{eigenpurposes} in the environment and associating them with options too \cite{machado2016learning}. Otherwise, surprise has been used as a criterion to build a curious agent exploring Atari games without reward with promising results \cite{burda2018large}. Other forms of reward have been designed in the intrinsic motivation literature, but usually to help explore and learn a single task in sparse rewards scenarios \cite{achiam2017surprise, bellemare2016unifying, jaderberg2016reinforcement}.

For the specific case of environment control, combining universal value function approximators (\uvfa) with goal conditioned policies (\gcp) has been proposed in the discrete action case and extended to the continuous case \cite{schaul2015universal, andrychowicz2017hindsight,ghosh2018learning}. In this case, the agent rewards itself when it reaches a goal state.

While this framework is natural for our agent objectives, it is not expressive enough to tackle object-level control. Indeed, goal-conditioned policies are limited to learn how to reach states from other states, but in an environment with multiple separately evolving, potentially non controllable objects, it is possibly detrimental and it makes limited sense to define control as reaching single states. For example, in a simple environment with two independent objects 1 and 2 characterized by their positions, the agent cannot use standard goal-conditioned policies to control object 2 without having to also set the position of object 1 to a specific value. Worse, if objects 1 and 2 states do not always transition independently, or if object 1 cannot be controlled, then some environment states are simply impossible to reach. In other words, the goals our agent needs to express to achieve the desired control must include reaching sets of states, those where one or several objects are in a desired configuration, independently from the others. To that end, we extend goal-conditioned policies to the control of specific features in a way described in Section \ref{sec:control}.

\subsection{Imitation learning for multiple skills}

The potential of learning from demonstrations \cite{Schaal_ANIPS_1997}(LfD) to accelerate skill acquisition is well-known in the context of \rl \cite{ijspeert2013dynamical,levine2013guided,hester2017deep}. In this work, we assume that: 1) the agent possibly acts independently of the learner, without the role of explicitly demonstrating useful behaviors, and 2) its actions have unknown intentions but result in coincidental demonstrations for various interactions with objects in the environment.

To our knowledge, very few works have tackled the issue of imitating third party agents that do not act as demonstrators. The focus has rather been on suboptimal demonstrations \cite{coates2008learning, argall2009survey} or identifying the best tutor among several ones \cite{duminy2019learning}.

A small number of works have proposed agents capable of imitating multiple skills from a tutor, either based on Bayesian inverse reinforcement learning with multiple intentions \cite{choi2012nonparametric, babes2011apprenticeship}, or using Generative Adversarial Networks to segment skills and imitate them simultaneously \cite{hausman2017multi}. These methods are difficult and heavy to adapt to the control problem described here, so we instead propose a simple heuristic to coarsely identify potential intentions and build upon DQN from demonstrations \cite{hester2017deep} to actually perform imitation. 

\subsection{Curriculum learning for reinforcement learning}

Curriculum learning has recently become the focus of intensive research in \rl \cite{graves2017automated,blaes2018control,narvekar2018learning,weinshall2018curriculum}. In the type of environment described, curriculum learning should be used to focus on promising objects and ignore those either too difficult or too easy for the current level of mastery of the agent. In other words, the agent needs to estimate its competence on controlling each object, and choose objects to train on so as to maximize its learning progress. 

Learning progress (\lp) maximization has already been used to learn multiple skills in an \rl context. Reinforcement learning can be used to sample skills to practice so as to maximize the expected global progress \cite{stout2010competence}. More often, task selection is framed as a non-stationary multi-armed bandit and \lp maximization is used as an effective heuristic to approach the problem \cite{baranes2010intrinsically}. In particular, \lp maximization has already been combined with standard goal-conditioned policies and deep reinforcement learning for a visual grid-world task, where goals were small raw-pixel observations \cite{veeriah2018many}. It proved relatively ineffective at outperforming random goal selection. Contrary to objects with different dynamics, the different goals in this setting share similarities and the network is likely to generalize much between them. Competence evaluations for goals thus may not remain up to date long enough to devise an effective learning progress measure. 

Our curriculum learning strategy relies on absolute \lp maximization \cite{baranes2010intrinsically}, and is especially close to that of the \curious algorithm \cite{colas2018curious}, where a multitask multigoal agent called \euvfa is also combined with absolute LP maximization to choose what task to train on. Using absolute values ensures the agent will refocus on a task for which its competence drops. The interplay of curriculum and imitation learning in an autonomous agent is present in several works which also demonstrate that imitation learning can drive the skill acquisition trajectory of an autonomous agent learning from intrinsic motivations \cite{nguyen2011bootstrapping,duminy2019learning}. However, these works address largely distinct technical issues and do not use deep \rl.

\section{Methods}
\label{methods}

\begin{figure}[t!]
\begin{minipage}[c]{.7\linewidth}
\includegraphics[width=\textwidth]{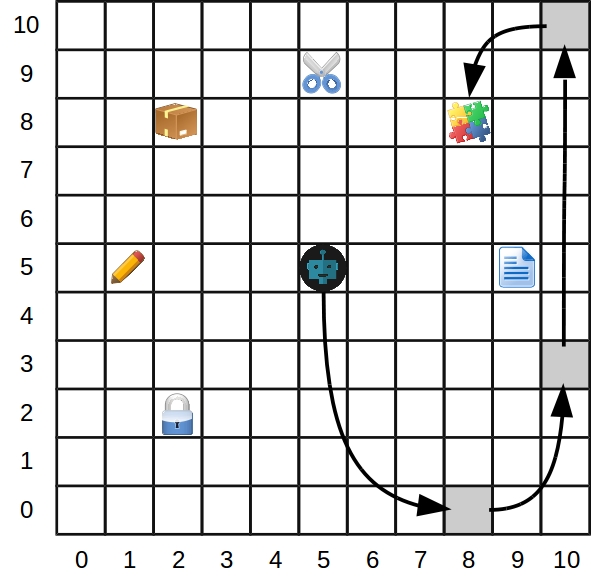}
\end{minipage}%
\hfill
\begin{minipage}[c]{.25\linewidth}
\includegraphics[width=\textwidth]{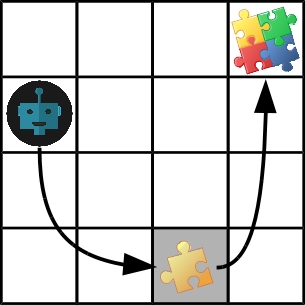}\\
\vfill
\includegraphics[width=\textwidth]{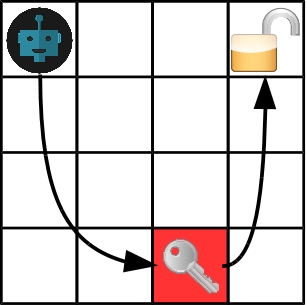}
\end{minipage}
\caption{\textbf{Left.} \esix, with six objects. To change the internal state of an object, the agent has to go through an unknown predefined ordered list of positions, shown here in gray for the jigsaw puzzle. These lists constitute an abstract representation of the necessity for the agent to go through specific intermediate states in order to control objects. \textbf{Top right.} Example of hierarchical relationship between two objects: the jigsaw piece is both an object and an intermediate position for the jigsaw puzzle, so that controlling the internal state of the puzzle requires controlling that of this individual piece. \textbf{Bottom right.} Example of partial control over features: one can imagine that Bob has a key that the agent does not have, and thus can open the lock and continue his actions, while the agent cannot: trying to reproduce Bob's actions in this case fails.}
\label{fig:environment}
\end{figure}

Our work involves the design of a discrete state, discrete action model of the environment, as described in the introduction, and the addition of a second unconstrained agent. We describe the model and the second agent in Section~\ref{sec:env} and Section~\ref{sec:bob}, while the specific instances used in the paper are described in Section~\ref{sec:instance} and Figure~\ref{fig:environment}. 

We then detail the \clic agent, which is based on three components: (1) it seeks object-level purposeful control of the environment by combining extended goal-conditioned policies and \ddqn (Section~\ref{sec:control}); (2) it observes and imitates the way an unconstrained agent Bob acts and changes individual objects states in the environment (Section~\ref{sec:imitation}); and (3) it builds and follows a curriculum over objects based on absolute learning progress maximization to choose what to learn and imitate (Section~\ref{sec:curri}). The outline of the full algorithm is given in Figure~\ref{alg1}.

\subsection{Environment model}
\label{sec:env}

In a realistic environment, stimuli associated to identical objects can only vary or be controlled together. Instead of perceiving the environment with raw sensors measurements and having to identify these objects from scratch, this work makes the simplifying assumption that the environment is readily perceived as a collection of separate objects, and focuses on learning how to control them independently and how to take advantage of the environment structure and of external agents to do so faster. 

The issue of identifying these environment features we call objects is arguably critical to build an agent that learns from scratch. Assuming an agent seeks to gain control of its environment, these features should be learned in a way that enables the agent to maximize this control, exactly as in traditional task-based deep \rl, where features learned from raw data are tailored to facilitate achieving a task. Some works have proposed to extract such features from raw data for independent control \cite{thomas2017independently}. However, this difficult topic lies outside the scope of this paper, and we keep this study and its combination with the present framework for future work.

The given objects in the environment may have various dynamics and be of distinct difficulties. To account for this, we define the environment as a grid world, and model each object in this world as a list of positions in the grid. For an object, these positions define intermediate stages in its dynamics. To illustrate this mechanism, take the example of a door: to open it, the agent should first grasp the handle, rotate it enough in some direction, and draw or pull. In a sense, this arbitrary decomposition of the action sequence needed to act on the door defines three via points to accomplish the desired control: "handle grasped", "handle rotated" and "door open". These via points would correspond to the first, second and third positions associated to the object "door", while the actions needed to reach each via point are modeled by moves in the grid to navigate between the three positions. The same way several moves can reach each via points, several trajectories in the grid can go through the three intermediate positions.

Formally, the environment is a $N \times N$ grid world containing $n$ objects $O_i$, where the agent can take five actions --- \textsc{right}, \textsc{left}, \textsc{up}, \textsc{down} and \textsc{act}. As explained above, each object $O_i$ is associated to a finite list of $l_i$ positions in the grid $L_i = (p_i^1, \cdots, p_i^{l_i})$. The object $O_i$ goes through several abstract internal states $o_i$ as the agent goes through the list of positions. Precisely $o_i$ takes its values in $\{0, \cdots, l_i\}$ and transitions only from $k$ to $k+1$ if the agent selects \textsc{act} on position $p_i^{k+1}$. The full environment state is the agent position in the grid plus the internal state of each object: $s_A = (x_A, y_A, o_1, \dots, o_n)$. 

This model enables to reproduce the environment properties we are interested in. For example, objects with close dynamics can simply have close or largely overlapping sequences of intermediate positions. An object more difficult to master is simply defined by a longer list of positions. Last, hierarchy between two objects $O_1$ and $O_2$ can be modeled by the inclusion of $L_1$ in $L_2$. For example in the previous case, one can argue that the handle and the door should be separate objects to maximize control, and that opening the door implies rotating the handle first (following positions in $L_{handle}$), and only then drawing the door properly to set open it: $L_{handle} \subset L_{door}$. 

\subsection{A second agent Bob}
\label{sec:bob}
To model the presence of other agents in the environment, a second agent called Bob can act in the grid, and \clic may imitate him. Every $F_{demo}$ steps, Bob achieves $d$ trajectories in the environment (Figure~\ref{alg1}, line 12), choosing an object $O_i$, a value $k \leq l_i$ and setting $O_i$ in state $k$. To obtain these trajectories, we rely on the property of our object model that the optimal sequence of actions to set object $i$ in state $k \in \{0, \cdots, l_i\}$ is defined by navigating through all $p_i^1$ to $p_i^k$ in the right order. To model objects controllable by Bob but not by the agent, some objects change internal states only when Bob goes through their list of positions, and not when the agent does so. Bob's choice of objects to act on depends on the experiment and is detailed for each experimental study.

\subsection{Environments description}
\label{sec:instance}

We carry out experiments in four $11\times 11$ grid worlds where objects can always be controlled by Bob, but not necessarily by \clic: 
\begin{itemize}
    \item \esix, with six objects controllable by \clic. Objects are independent and of the same difficulty: $L_i \cap L_j = \varnothing$, and $l_i = l_j$ for all $i,j$.
    \item \eun, \etrois, two partially controllable variations of \esix with the same objects, but where \clic can control respectively one and three of the six objects.
    \item \eh with six fully controllable but hierarchically related objects: $L_1 \subset L_2 \subset \cdots \subset L_6$. In this case, objects 4, 5 and 6 are more difficult to master than objects in \esix, while objects 1 and 2 are easier.
\end{itemize}

Both \clic and Bob always start in the center of the grid, and stochasticity is introduced in the environment through a {\em sticky action} phenomenon: they repeat their last action with probability $0.25$ instead of following their policy \cite{machado2017revisiting}. \clic perceives and stores the states visited by Bob as $s_{Bob} = (x_{Bob}, y_{Bob}, o_1, \dots, o_6)$, and we assume that the only source of non-optimality of Bob's actions comes from sticky actions.

\subsection{Separate control of objects}
\label{sec:control}

Let $\mathcal{M}$ be a Markov Decision Processes without reward function, with state space $\mathcal{S}$, and discrete action space $\mathcal{A}$. With standard goal-conditioned policies, the agent learns to reach $\bm{g} = (g_1, \dots, g_n)$ from $\bm{s} = (s_1, \dots, s_n)$, by maximizing $R_{\bm{g}}$, where $R_{\bm{g}}(s) = 0$ if $\norm*{\bm{g} - \bm{s} }\leq \epsilon$, $-1$ otherwise. This parameterization does not let the agent focus on certain aspects of the environment and ignore others; instead it is forced to set $s_i = g_i$ for all $i$. 

To alleviate this restriction, we modify the reward function family to weight each feature $s_i$ and write
\begin{equation}
\label{eq:reward}
R_{\bm{g}, \bm{w}}(\bm{s})=
\begin{cases}
  0, & \text{if}\ \abs*{\bm{w} \cdot (\bm{g} - \bm{s}) }\leq \epsilon \\
  -1, & \text{otherwise}
\end{cases}
\end{equation}
where $\cdot$ denotes the dot-product and $\bm{w} = (w_1, \dots, w_n)$ is a normalized weight vector with values between 0 and 1. It describes how much setting each feature to its goal value matters to the agent when it comes to rewarding itself. For example if $w_1 = w_2 = 1$ and all others $w_i$ are zeros, then the agent is aiming at any state where $s_1 = g_1$ and $s_2 = g_2$, independently of other $s_i$.

In our gridworld environments, we are interested in the simplified objective of individual object control only, and the agent state writes $s_A = (x_A, y_A, o_1, \dots, o_n)$. In this case, for each object $O_i$, we define $w^i$ the one-hot encoded vector with a one a the position of $o_i$ in $s_A$ and zeros elsewhere. Then for a goal \emph{value} $g \in \{0, \dots, l_i\}$, we write $R_{g, \bm{w^i}} = 0$ if $\norm*{g - \bm{w^i \cdot \bm{s}} }\leq \epsilon$, $-1$ otherwise. These reward functions indeed correspond to independent control of objects states. 

While the family of reward functions defined by Equation~\ref{eq:reward} enables to work with any weight vectors and control any number of objects in the environment, we keep the study of this multi-object control for future work. Instead, at the start of each episode the agent follows its curriculum to select an object $O_i$ to practice on (Section~\ref{sec:curri}, Figure~\ref{alg1}, line 2 and 9), a random goal value $g$ between 1 and $l_i$, and acts to maximize $R_{g, \bm{w^i}}$. 

To this end, \clic uses \uvfas and approximates the object-related action-value function $Q_{\pi}^{g, \bm{w^i}}(\bm{s}, a)$ with a neural network, taking $\bm{s}$, $\bm{w^i}$ and $g$ as inputs. The agent acts following a softmax exploration strategy and stores the transitions in a standard replay buffer. At each step it minimizes the double \dqn loss $J_{DQ}$ \cite{van2015deep}:
\begin{multline}
J_{DQ}^{g, \bm{w^i}}(Q) = \big[R_{g, \bm{w^i}}(\bm{s'}) + \gamma \tilde{Q}^{g, \bm{w^i}}(\bm{s'}, \argmaxG_{a}Q^{g, \bm{w^i}}(\bm{s'}, a)) \\- Q^{g, \bm{w^i}}(\bm{s}, a)\big]^2 
\label{eq1}
\end{multline} 
where $\tilde{Q}$ is the target network.


\subsection{Imitation}
\label{sec:imitation}

As explained in the introduction, a single trajectory from Bob may contain fortuitous executions of several control objectives. In this work, \clic simply observes and tries to reproduce Bob's actions that change the states of objects it seeks to control, whether these changes were the true goals of the actions or not. Precisely, if Bob changed and set object state $o_i$ to value $g$ at some point, then Bob's actions up to this point constitute a demonstration --- not necessarily optimal --- for setting $o_i$ to $g$, or in other words to maximize $R_{g,\bm{w_i}}$.

Each transition $(\bm{s_B}, a_B, \bm{s'_B})$ from the $d$ instances of Bob trajectories is augmented with $(g, \bm{w^i})$ pairs identified this way, and \clic can compute the associated rewards at training time for all such pairs. The augmented transitions are stored both in the same replay buffer as the agent's own experience (Figure~\ref{alg1}, line 15), and in a separate set $D$ for imitation learning (line 12).

\begin{figure}
\caption{\clic: Curriculum Learning and Imitation for Control}
\label{alg1}
\begin{algorithmic}[1]
\State {\bfseries Input:} transition function $T$, empty replay buffer $RB$
\State \textbf{Initialize:} state $s$, $O_i \sim p_{k, \epsilon}$, $g \sim \mathcal{U}(\{0, \dots, l_i\})$, step $k=0$
\Loop
\State $a \sim softmax_a Q^{g, \bm{w^i}}(\bm{s},a)$
\State $s' \sim T(\bm{s},a)$
\State $RB \leftarrow (\bm{s}, a, s', g, \bm{w^i})$
\State Minimize \eqref{eq1} on batch from $RB$
\If{terminal or timeout}
\State $O_i \sim p_{k, \epsilon}$, $g \sim \mathcal{U}(\{0, \dots, l_i\})$
\EndIf
\If{k \% $F_{demo}$ = 0}
\State $D \leftarrow (\bm{s_B}, a_B, \bm{s'_B})$\Comment{Bob's $d$ demonstrations}
\ForAll{$(\bm{s_B}, a_B, \bm{s'_B}) \in D$}
\State Augment $(\bm{s_B}, a_B, \bm{s'_B})$ with $(g, \bm{w^i})$\Comment{\ref{sec:imitation}}
\State $RB \leftarrow (\bm{s_B}, a_B, \bm{s'_B}, g, \bm{w^i})$
\EndFor
\For{$N_{imit}$ steps}
\State $O_i \sim p_{k, \epsilon}$
\State Sample batch $B$ of $(\bm{s_B}, a_B, \bm{\bm{s'_B}}, g, \bm{w^i}) \in D$
\State Minimize $\eqref{eq1} + \eqref{eq2}$ on $B$
\EndFor
\State Empty $D$
\EndIf
\State $k=k+1$
\EndLoop
\end{algorithmic}
\end{figure}
 
After trajectories from Bob are observed, the agent imitates him for $N_{imit}$ steps (Figure~\ref{alg1}, lines 18-20). Adapting DQNfD \cite{hester2017deep} to the multi-goal case, we define a large margin classification loss for our general Q-values $J_{I}^{g, \bm{w^i}}(Q)$ that writes
\begin{multline}
J_{I}^{g, \bm{w^i}}(Q) = \max_{a \in \mathcal{A}}\left[Q^{g, \bm{w^i}}(\bm{s_B}, a) + l(a_B, a)\right] \\- Q^{g, \bm{w^i}}(\bm{s_B}, a_B)
\label{eq2}
\end{multline}
where $l(a_B,a)$ is a margin function that is $0$ when $a=a_B$ and $1$ otherwise. This loss ensures that the values for the agent of Bob's actions will be at least a margin above Q-values of other actions.

At each of the $N_{imit}$ imitation step, \clic selects an object following Section~\ref{sec:curri}, and minimizes $J_{DQ} + J_{I}$ on batches of transitions 
observed for this object.

\subsection{Curriculum learning}
\label{sec:curri}

\clic uses its own learning progress to select which object to practice on (Section~\ref{sec:control}, Figure~\ref{alg1}, lines 2, 9), and which object to imitate Bob on (Section~\ref{sec:imitation}, Figure~\ref{alg1}, line 18). To that end, it tracks its learning progress on each object of interest, and samples preferably those for which this learning progress is maximal. 

Similar to \curious \cite{colas2018curious}, we define the agent's competence $C_k(O_i) \in [0, 1]$ at step $k$ for object $O_i$ as the average success over a window of $l$ tentative episodes at controlling $O$ --- success meaning $R_{\bm{g}, \bm{w_i}}(s) = 0$ at the end of episode, with desired value $g$ for $o_i$. The learning progress then writes
$LP_k(O_i) = \abs*{C_k(O_i) - C_{k-l}(O_i)}$. It is used to derive for each object a sampling probability at step $k$:
$$
p_{k,\epsilon}(O_i) = \epsilon \times \frac{1}{N} + (1 - \epsilon) \times \frac{LP_k(O_i)}{\sum_{f} LP_k(O_i)}
$$
with $\epsilon$ controlling the sampling randomness: $\epsilon=1$ means pure random sampling, while $\epsilon=0$ means pure proportional sampling. To ensure a minimum amount of exploration of objects, \clic uses $\epsilon=0.2$. We call \clicrnd the version of \clic that samples objects randomly without maximizing \lp, using $\epsilon=1$. The use of absolute values in \lp ensures a competence drop on a feature leads to training again on this feature \cite{baranes2013active,colas2018curious}.

\subsection{Summary and parameters}

In summary, the agent starts an episode by sampling an object to act upon and a goal value to set its state to. \clicrnd performs this sampling randomly while \clic chooses an object that it is likely to control better after practicing on it. From there, it alternates phases of autonomous learning using the \ddqn algorithm with phases of imitation learning using the \dqnfd algorithm, both extended to the multi-goal setting. When observing Bob's actions, \clic simply stores the transitions that lead to objects states changes, to train on them later.

All \clic parameters are as follows: we take $N_{imit} = 100$, $F_{demo} = 5000$, $d=25$, $l=100$, an episode timeout is reached after 200 steps, the batch size is 64, the temperature for the softmax exploration is $1$ and the replay buffer size is $1e^5$. Q-values are approximated with a neural network with two 32-neurons hidden layers with {\sc relu} activations \footnote{The code of the experiments is available at \url{https://www.dropbox.com/s/zs1ukyo0la7hze2/clic.zip?dl=0}.}.

\section{Results}
\label{sec:results}

In all figures, $C(k) = \sum_i C_k(O_i) / N_c$ is the average normalized competence at step $k$, with $N_c$ the number of controllable objects, and $I_k(O_i)$ is the number of steps \clic spends imitating Bob on $O_i$ (max. $N_{imit} = 100$). Unless stated otherwise, when Bob is said to act on a set of objects, he samples randomly one of these object $O_i$ at the start of all his $d$ trajectories and acts to set $o_i = 1$. Plain curves correspond to the median of 10 runs and shaded areas to their interquartile ranges. 

In these experiments we examine: 1) if \clic can identify useful demonstrations for how to change object states in Bob's actions, independently of his true intentions; 2) to what extent \clic's learning trajectory can be influenced by Bob's behaviors; and 3) if \clic can identify and ignore behaviors that should not be reproduced, because it will not be able to do so, or already knows how to do so. As explained in Section~\ref{sec:curri}, the curriculum learning agent \clic uses proportional sampling with $\epsilon = 0.2$ whereas the basic agent \clicrnd uses random sampling with $\epsilon = 1$. 

\subsection{Imitating Bob}

\begin{figure}[!t]
\includegraphics[width=\linewidth]{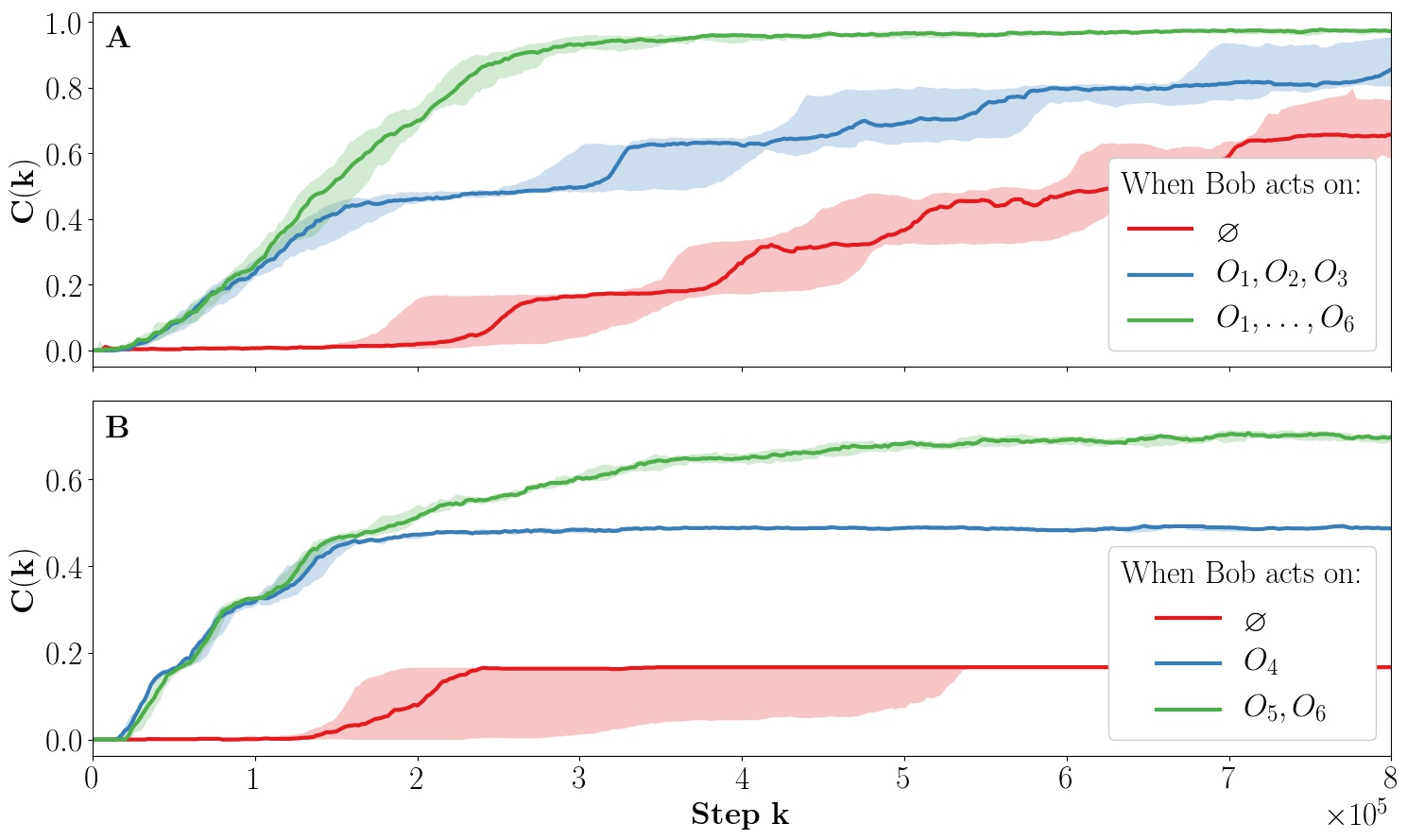}
\caption{\textbf{A}. Average competence of \clic in \esix, when Bob does nothing or controls three or six objects. \textbf{B}. Average competence of \clic agent in \eh, when Bob does nothing or controls one intermediate object or the two hardest of the hierarchy.}
\label{fig:result1}
\end{figure}

First we analyze the impact of observing and imitating Bob, both in \esix (\figurename~\ref{fig:result1} A) and \eh (\figurename~\ref{fig:result1} B). In \esix, Bob either does nothing, or acts on objects 1, 2 and 3, or acts on all objects. It is clear that \clic learns faster when it sees Bob acting on more objects. 

In \eh, Bob either does nothing, or controls the intermediate object 4 or the two hardest of the hierarchy, 5 and 6. This environment is harder to control in autonomy as it requires more exploration to discover the hard objects. Results show that \clic uses Bob's behavior to gain control over the objects Bob seeks to control, but also over those Bob controls as intermediate steps. 

For example, when Bob controls Objects 5 and 6, the hierarchy implies that he has to also act on all other objects before. In this case, results show that \textsc{\textbf{clic}} \textbf{can imitate Bob's actions to control more than only what Bob intended to achieve.} When Bob controls only Object 4, \clic gains control over objects 1 to 4 as well, but discovering autonomously harder ones is too difficult (the lists of positions that define them is too long).

\subsection{Following Bob's teaching}
\label{sec:guide}

In \figurename~\ref{fig:result1} A, when Bob controls only a subset of \esix objects, \clic first imitates him to reach $C \approx 0.5$, and then explores the rest of the environment autonomously. From another point of view, these results prove that Bob can influence \clic's learning trajectory by only showing it how to control some selected objects. We now show that, acting as a mentor, Bob can completely control this learning trajectory.

\begin{figure}[t!]
\includegraphics[width=\linewidth]{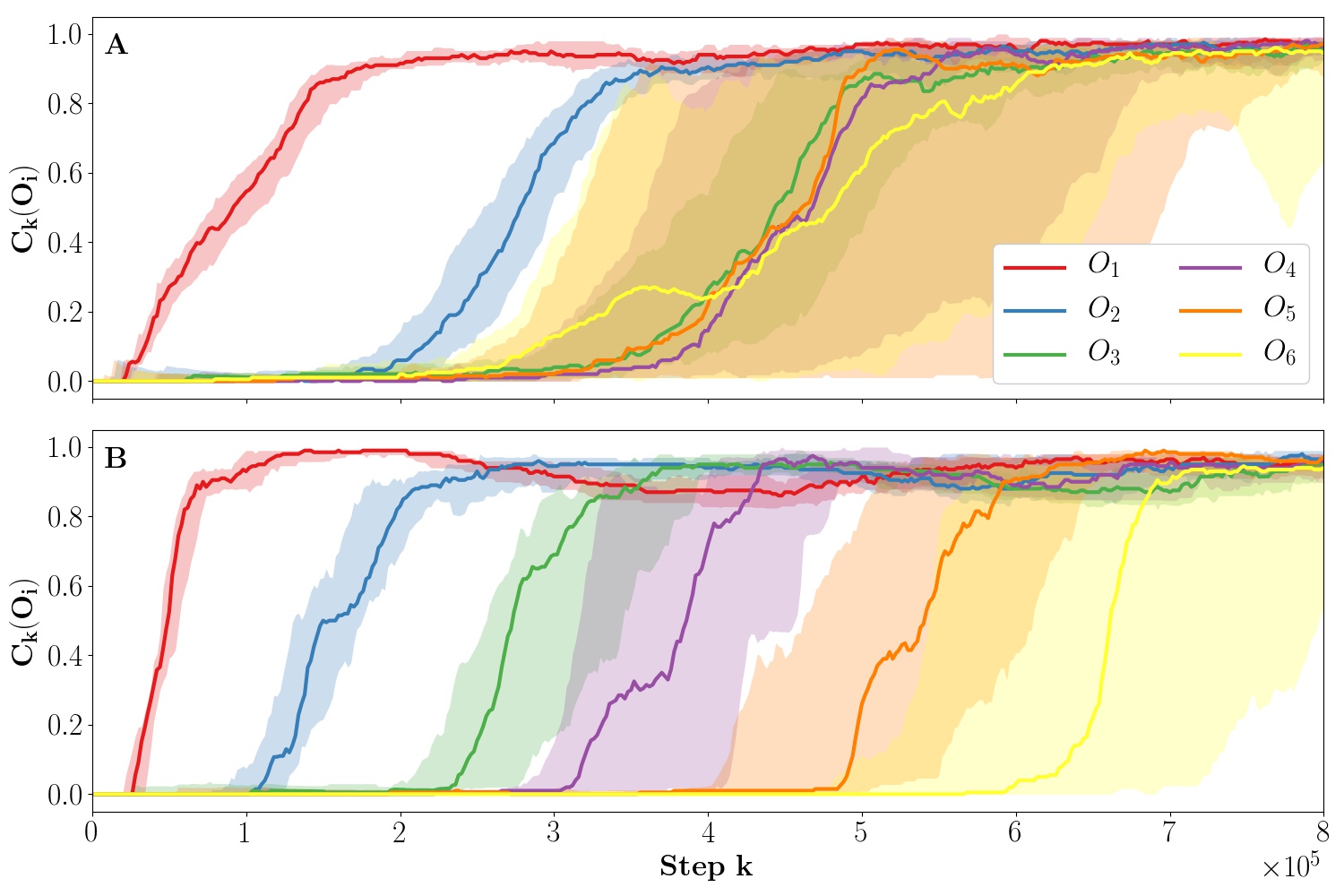}
\caption{Competence for each object in \esix, when Bob provides demonstrations in order, for \clicrnd (\textbf{A}) and \clic (\textbf{B}). LP maximization enables the agent to follow Bob's demonstrations order.}
\label{fig:result2}
\end{figure}

In \figurename~\ref{fig:result2}, instead of choosing randomly an object to control, Bob shows how to control Object 1 until the agent's competence for it is above 0.9. Then Bob demonstrates control over Object 2, and so on. If the agent's competence on a previously mastered object falls below 0.9, Bob provides demonstrations for it again. The figure compares the performances of \clic (\figurename~\ref{fig:result2} B) and \clicrnd (\figurename~\ref{fig:result2} A), in \esix (where objects are of the exact same difficulties) and with Bob acting as specified.

Without curriculum learning, \clicrnd's learning trajectory starts to follow Bob's demonstrations order but does not do so up to the end: as \clicrnd learns autonomously in parallel of imitating Bob, and randomly chooses what to learn in autonomy, nothing prevents it from learning to control objects 4, 5 and 6 before Bob shows how to do so.

By contrast, \clic strictly follows the curriculum taught by Bob, learning almost always one object at a time. This time, during its autonomous learning phase, LP maximization pushes \clic to stick to the objects Bob demonstrated, until it does not make progress on them anymore. In other words, \textbf{\textsc{clic} can be taught control over the environment in a desired order by simply showing it demonstrations in this order.}

We observe that following Bob's order does not speed up global control of the environment. In \esix, all objects are independent, so there is no transfer between learning the different objects. The ability to focus on learning them one by one is balanced by the fact that the network overfits when it trains on one only at a time.

\subsection{Ignoring non-reproducible behaviors from Bob}

\begin{figure}[!t]
\includegraphics[width=\linewidth]{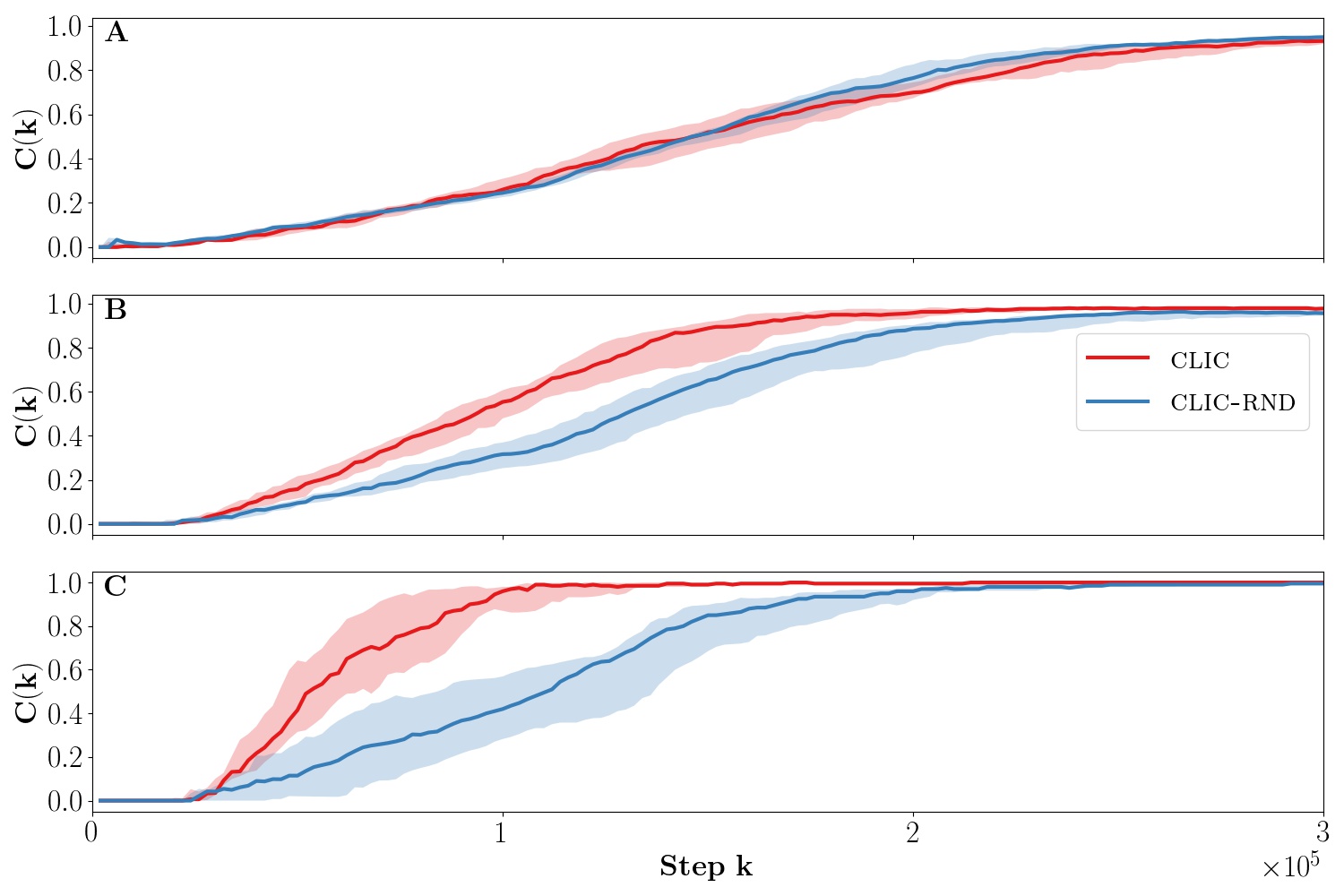}
\caption{Average competence of \clicrnd and \clic over all controllable objects in \textbf{A}. \esix, \textbf{B}. \etrois and \textbf{C}. \eun. The benefits of LP maximization increase with the number of uncontrollable aspects of the environment.}
\label{fig:result3}
\end{figure}

When Bob controls objects that the agent cannot, imitating Bob can be a waste of time. Yet, if the agent knew in advance the subset of objects it cannot control and imitate Bob on, learning would be simpler as the agent could focus on fewer aspects of the environment. But in this work, \clic does not have this knowledge and relies on curriculum learning to discover what to learn and imitate.

In \figurename~\ref{fig:result3}, we perform experiments in (A) $E_6$ with all objects controllable by the agent, (B) $E_3$ with half of them controllable, and (C) $E_1$ with only one controllable. In each environment, Bob acts on all objects, including those that the agent has no control on, and we study the impact of LP maximization on the agent performances. 

Comparing the three blue curves shows that, without curriculum learning, having fewer objects to master boosts only slightly \clicrnd performances on global control. As expected, without a mechanism to tell between what can and cannot be learned, much of the advantage of having fewer to learn is lost. Also, we note that the variability of \clicrnd  increases with the number of uncontrollable objects: achieving global control becomes dependent on the probability that it discovers the only controllable objects more or less early.

In $E_6$, when all objects are controllable, LP maximization does not bring any benefit and \clic does not outperform \clicrnd. Indeed objects are independent and of equal difficulties, so there can only be a poor form of transfer between them, and a low gain from practicing them and imitating Bob on them in any specific order, as LP maximization pushes the agent to do. 

In $E_1$ and $E_3$, on the contrary, the red curves show that \textbf{LP maximization helps achieving maximum control faster in presence of uncontrollable objects}: when three or five objects are uncontrollable, \clic learns faster than \clicrnd. Besides, the impact of curriculum learning is greater when there are more uncontrollable objects: the gap between \clicrnd and \clic performances is wider in $E_1$ than it is in $E_3$.

\begin{figure}[!t]
\includegraphics[width=\linewidth]{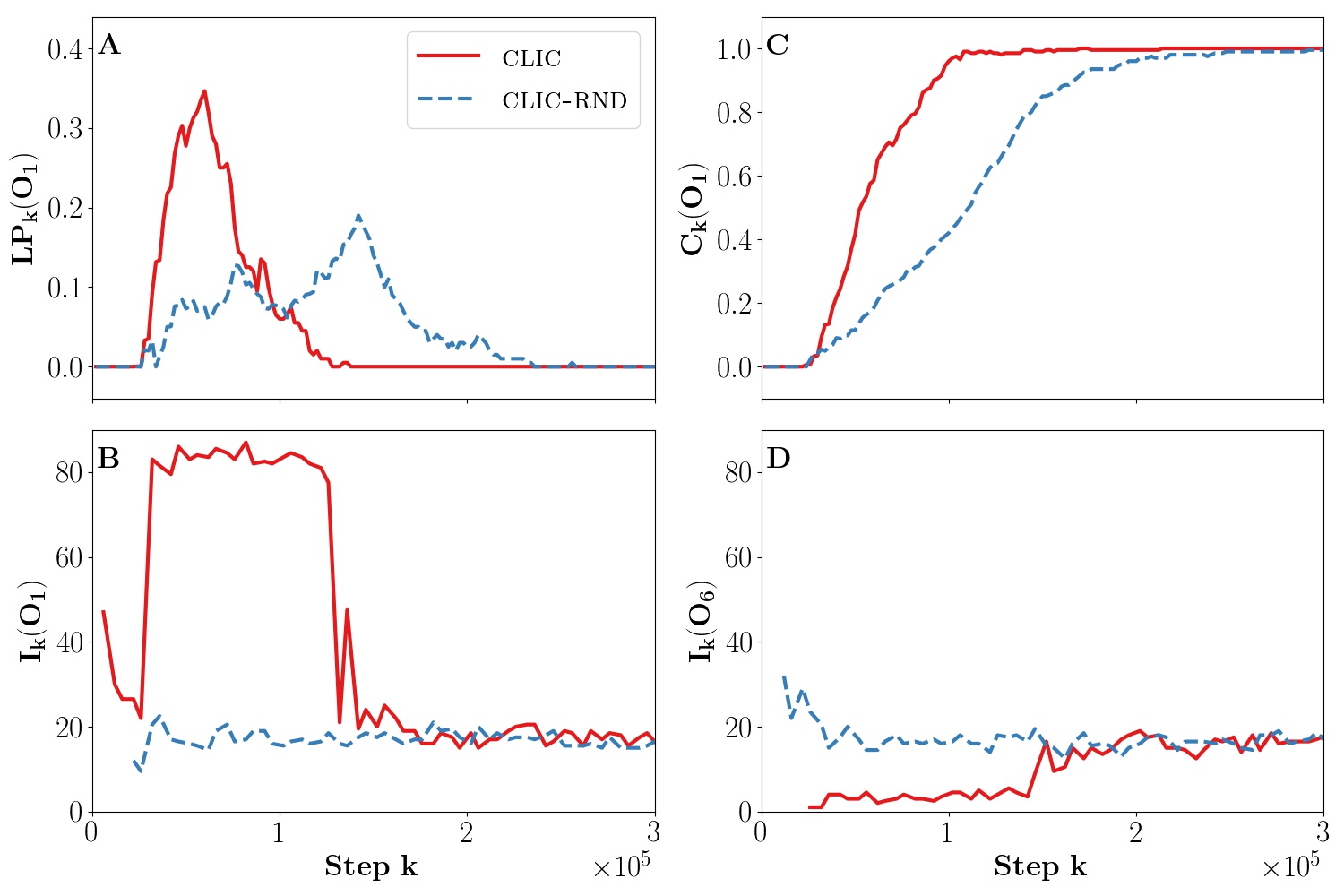}
\caption{Effect of LP maximization in environment \eun with one controllable object $O_1$: \textbf{A}. Learning progress $LP_k(O_1)$; \textbf{B}. Steps spent imitating Bob controlling $O_1$, $I_k(O_1)$; \textbf{C}. Competence $C_k(O_1)$ (same as \figurename~\ref{fig:result1}.c); \textbf{D}. $I_k(O_6)$ with $O_6$ not controllable by \clic.}
\label{fig:result4}
\end{figure}

Reasons for this gain are found in \figurename~\ref{fig:result4}, which shows the impact of LP maximization when only $O_1$ is controllable. \figurename~\ref{fig:result4} displays the evolution of the agent learning progress (A) and competence (B) on $O_1$, the amount of steps spent imitating Bob controlling $O_1$ (C), and the amount of steps uselessly spent imitating Bob controlling $O_6$ (D).

When it does not maximize LP (dashed blue), the agent ignores the LP bump (\figurename~\ref{fig:result4} A) resulting from gaining some control over the object (\figurename~\ref{fig:result4} C). So it randomly selects what to train on, and above all what to imitate: the agent tries to reproduce Bob's behavior when it can (\figurename~\ref{fig:result4} B) as often as when it cannot (\figurename~\ref{fig:result4} D), as shown by the blue curves at the same level. This is sub-optimal as performances on uncontrollable objects will never increase. 

Instead, when the agent maximizes LP (plain red), the bump in LP at 50k steps (\figurename~\ref{fig:result4} A) results in more imitation for this object (\figurename~\ref{fig:result4} B) and more episodes trying to control it. Simultaneously, the agent stops focusing on other objects for which its learning progress is too small, among which uncontrollable ones (\figurename~\ref{fig:result4} D). This focus results in a faster learning pace on controllable objects (\figurename~\ref{fig:result4} C).

\subsection{Ignoring Bob's demonstrations for mastered objects}

By contrast with \eun to \esix, in \eh, all objects are controllable but some are easier than others. In particular, controlling Object $i$ is easier than controlling Object $i+1$, and any demonstration for Object $i+1$ contains a demonstration for Object $i$. 

As a consequence, in such an environment, when Bob acts on random objects, he ends up providing more demonstrations for the easiest objects. If Bob chooses to guide the agent as in Section~\ref{sec:guide}, he initially focuses on these easy objects, and the bias towards them is even stronger.  

\begin{figure}[!t]
\includegraphics[width=\linewidth]{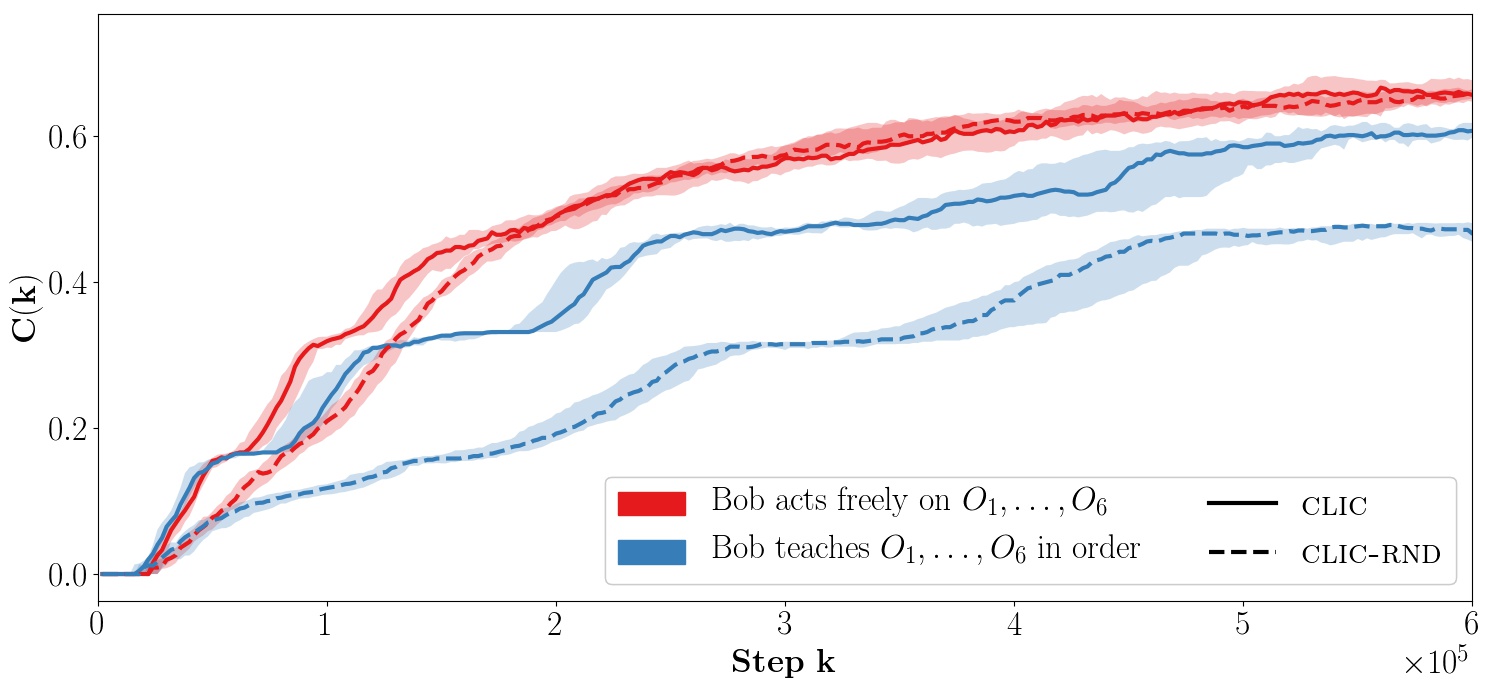}
\caption{Average competence of \clicrnd and \clic in $E_h$, depending on Bob's policy: selecting randomly objects to control, or teaching the agent in order.}
\label{fig:result5}
\end{figure}

\figurename~\ref{fig:result5} shows the global performances of \clicrnd and \clic in these two scenarios where actions from Bob are biased towards easy objects. In both scenarios, \textbf{curriculum learning helps ignoring Bob's behavior affecting already mastered objects.} The effect of curriculum is greater when Bob guides the agent, as the bias is stronger in this case. But the same effect holds when Bob chooses randomly what to do, at least early in learning.

\begin{figure}[!t]
\includegraphics[width=\linewidth]{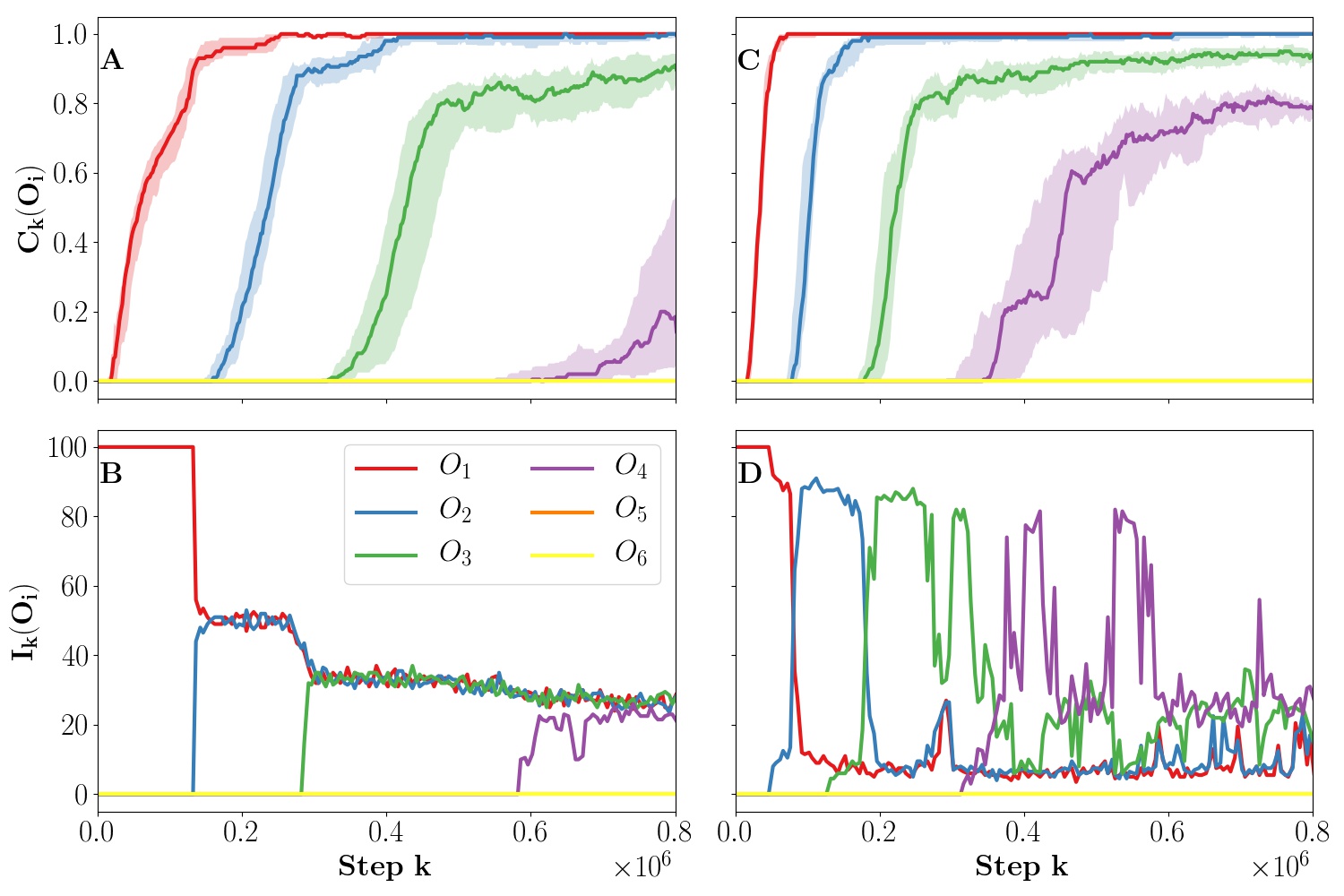}
\caption{In \eh, when Bob teaches objects $O_1$ to $O_6$ in order: \textbf{A}. Competence of \clicrnd on each object. \textbf{B}. Steps spent by \clicrnd imitating Bob controlling each object. \textbf{C}. Competence of \clic on each object. \textbf{D}. Steps spent by \clic imitating Bob controlling each object.}
\label{fig:result6}
\end{figure}

The fact that curriculum learning enables \clic to stop imitating Bob on easy objects can be clearly seen in \figurename~\ref{fig:result6}. Here Bob guides the agent, and the bias towards easy objects is strong. \clicrnd chooses randomly what to imitate among what it is shown, so it imitates Bob mainly on quickly mastered objects. 

For instance, at step 400K, Object 3 is not yet mastered by \clicrnd so Bob demonstrates it, along with its necessary intermediate steps Objects 1 and 2; \clicrnd, observing these three objects being controlled, imitates Bob on all of them (\figurename~\ref{fig:result6} B), whereas it already knows how to control Objects 1 and 2. Thus it loses precious imitation steps to make progress on Object 3.

Instead, thanks to LP maximization, \clic stops imitating Bob on Object 1 (\figurename~\ref{fig:result6} D) as soon as its competence on it reaches 1 and stops rising (\figurename~\ref{fig:result6} C). This mechanism enables \clic to learn more and faster by focusing its learning resources on non-mastered objects only. 

\section{Conclusion}
\label{sec:conclu}

In this work, we proposed a new learning setting, with a non-rewarding environment where a third party agent called Bob acts without communicating the intent of its actions and in ways that can be non-reproducible for the agent. 
This setting, although discrete, is a first step towards real-life environments, where artificial agents will face a very large number of potential tasks and goals, next to other agents with different capabilities.
We combined object control, curriculum and imitation learning to build an agent called \clic that addresses the issues raised by this challenging learning context.

In particular, we showed that \clic predictably makes faster progress when it observes more behaviors from Bob, but also that it can leverage Bob's actions to make progress for more tasks than only those demonstrated. We demonstrated that Bob could mentor \clic and control its developmental trajectory by simply providing ordered demonstrations. Eventually, we showed that \clic can effectively use learning progress maximization to tell between what is and is not useful to learn and imitate, and thus learns faster both when the environment is partially controllable and when it contains a natural hierarchy.

A specificity of this work was that, rather than considering a human expert teaching the agent, as is usually the case in interactive learning, we considered an external artificial agent providing non intentional demonstrations.
As a consequence, we did not focus on limiting the amount of demonstrated behaviors, as is often the case in the domain \cite{kang2018policy}. Another topic of interest that emerges from our work is the importance of the order of the demonstrations performed by a teaching agent. We keep these two topics for future work.

Additionally, our focus being on the combination of curriculum learning and imitation learning rather than on representation learning, experiments were performed in discrete state, discrete action environments with independent objects. But the main ideas presented here could easily be extended to the continuous case, replacing \dqnfd with \ddpgfd \cite{vevcerik2017leveraging} and trying to learn independently controllable objects \cite{thomas2017independently}.
 
\section*{Acknowledgment}
Anonymized.

\bibliographystyle{IEEEtran}
\bibliography{local}

\begin{thebibliography}{10}
\providecommand{\url}[1]{#1}
\csname url@samestyle\endcsname
\providecommand{\newblock}{\relax}
\providecommand{\bibinfo}[2]{#2}
\providecommand{\BIBentrySTDinterwordspacing}{\spaceskip=0pt\relax}
\providecommand{\BIBentryALTinterwordstretchfactor}{4}
\providecommand{\BIBentryALTinterwordspacing}{\spaceskip=\fontdimen2\font plus
\BIBentryALTinterwordstretchfactor\fontdimen3\font minus
  \fontdimen4\font\relax}
\providecommand{\BIBforeignlanguage}[2]{{%
\expandafter\ifx\csname l@#1\endcsname\relax
\typeout{** WARNING: IEEEtran.bst: No hyphenation pattern has been}%
\typeout{** loaded for the language `#1'. Using the pattern for}%
\typeout{** the default language instead.}%
\else
\language=\csname l@#1\endcsname
\fi
#2}}
\providecommand{\BIBdecl}{\relax}
\BIBdecl

\bibitem{Schaal_ANIPS_1997}
S.~Schaal, ``Learning from demonstration,'' in \emph{Advances in Neural
  Information Processing Systems 9}.\hskip 1em plus 0.5em minus 0.4em\relax
  Cambridge, MA: MIT Press, 1997, pp. 1040--1046.

\bibitem{eysenbach2018diversity}
B.~Eysenbach, A.~Gupta, J.~Ibarz, and S.~Levine, ``Diversity is all you need:
  Learning skills without a reward function,'' \emph{arXiv preprint
  arXiv:1802.06070}, 2018.

\bibitem{gregor2016variational}
K.~Gregor, D.~J. Rezende, and D.~Wierstra, ``Variational intrinsic control,''
  \emph{arXiv preprint arXiv:1611.07507}, 2016.

\bibitem{warde2018unsupervised}
D.~Warde-Farley, T.~Van~de Wiele, T.~Kulkarni, C.~Ionescu, S.~Hansen, and
  V.~Mnih, ``Unsupervised control through non-parametric discriminative
  rewards,'' \emph{arXiv preprint arXiv:1811.11359}, 2018.

\bibitem{machado2016learning}
M.~C. Machado and M.~Bowling, ``Learning purposeful behaviour in the absence of
  rewards,'' \emph{arXiv preprint arXiv:1605.07700}, 2016.

\bibitem{burda2018large}
Y.~Burda, H.~Edwards, D.~Pathak, A.~Storkey, T.~Darrell, and A.~A. Efros,
  ``Large-scale study of curiosity-driven learning,'' \emph{arXiv preprint
  arXiv:1808.04355}, 2018.

\bibitem{achiam2017surprise}
J.~Achiam and S.~Sastry, ``Surprise-based intrinsic motivation for deep
  reinforcement learning,'' \emph{arXiv preprint arXiv:1703.01732}, 2017.

\bibitem{bellemare2016unifying}
M.~Bellemare, S.~Srinivasan, G.~Ostrovski, T.~Schaul, D.~Saxton, and R.~Munos,
  ``Unifying count-based exploration and intrinsic motivation,'' in
  \emph{Advances in Neural Information Processing Systems}, 2016, pp.
  1471--1479.

\bibitem{jaderberg2016reinforcement}
M.~Jaderberg, V.~Mnih, W.~M. Czarnecki, T.~Schaul, J.~Z. Leibo, D.~Silver, and
  K.~Kavukcuoglu, ``Reinforcement learning with unsupervised auxiliary tasks,''
  \emph{arXiv preprint arXiv:1611.05397}, 2016.

\bibitem{schaul2015universal}
T.~Schaul, D.~Horgan, K.~Gregor, and D.~Silver, ``Universal value function
  approximators,'' in \emph{International Conference on Machine Learning},
  2015, pp. 1312--1320.

\bibitem{andrychowicz2017hindsight}
M.~Andrychowicz, F.~Wolski, A.~Ray, J.~Schneider, R.~Fong, P.~Welinder,
  B.~McGrew, J.~Tobin, P.~Abbeel, and W.~Zaremba, ``Hindsight experience
  replay,'' \emph{arXiv preprint arXiv:1707.01495}, 2017.

\bibitem{ghosh2018learning}
D.~Ghosh, A.~Gupta, and S.~Levine, ``Learning actionable representations with
  goal-conditioned policies,'' \emph{arXiv preprint arXiv:1811.07819}, 2018.

\bibitem{ijspeert2013dynamical}
A.~J. Ijspeert, J.~Nakanishi, H.~Hoffmann, P.~Pastor, and S.~Schaal,
  ``Dynamical movement primitives: learning attractor models for motor
  behaviors,'' \emph{Neural computation}, vol.~25, no.~2, pp. 328--373, 2013.

\bibitem{levine2013guided}
S.~Levine and V.~Koltun, ``Guided policy search,'' in \emph{Proceedings of the
  30th International Conference on Machine Learning}, 2013, pp. 1--9.

\bibitem{hester2017deep}
T.~Hester, M.~Ve{\v{c}}er{\'\i}k, O.~Pietquin, M.~Lanctot, T.~Schaul, B.~Piot,
  D.~Horgan, J.~Quan, A.~Sendonaris, G.~Dulac-Arnold \emph{et~al.}, ``Deep
  q-learning from demonstrations,'' \emph{arXiv preprint arXiv:1704.03732},
  2017.

\bibitem{coates2008learning}
A.~Coates, P.~Abbeel, and A.~Y. Ng, ``Learning for control from multiple
  demonstrations,'' in \emph{Proceedings of the 25th international conference
  on Machine learning}.\hskip 1em plus 0.5em minus 0.4em\relax ACM, 2008, pp.
  144--151.

\bibitem{argall2009survey}
B.~D. Argall, S.~Chernova, M.~Veloso, and B.~Browning, ``A survey of robot
  learning from demonstration,'' \emph{Robotics and autonomous systems},
  vol.~57, no.~5, pp. 469--483, 2009.

\bibitem{duminy2019learning}
N.~Duminy, S.~M. Nguyen, and D.~Duhaut, ``Learning a set of interrelated tasks
  by using sequences of motor policies for a socially guided intrinsically
  motivated learner,'' \emph{Frontiers in Neurorobotics}, 2019.

\bibitem{choi2012nonparametric}
J.~Choi and K.-E. Kim, ``Nonparametric bayesian inverse reinforcement learning
  for multiple reward functions,'' in \emph{Advances in Neural Information
  Processing Systems}, 2012, pp. 305--313.

\bibitem{babes2011apprenticeship}
M.~Babes, V.~Marivate, K.~Subramanian, and M.~L. Littman, ``Apprenticeship
  learning about multiple intentions,'' in \emph{Proceedings of the 28th
  International Conference on Machine Learning (ICML-11)}, 2011, pp. 897--904.

\bibitem{hausman2017multi}
K.~Hausman, Y.~Chebotar, S.~Schaal, G.~Sukhatme, and J.~J. Lim, ``Multi-modal
  imitation learning from unstructured demonstrations using generative
  adversarial nets,'' in \emph{Advances in Neural Information Processing
  Systems}, 2017, pp. 1235--1245.

\bibitem{graves2017automated}
A.~Graves, M.~G. Bellemare, J.~Menick, R.~Munos, and K.~Kavukcuoglu,
  ``Automated curriculum learning for neural networks,'' \emph{arXiv preprint
  arXiv:1704.03003}, 2017.

\bibitem{blaes2018control}
S.~Blaes, M.~Vlastelica, J.-J. Zhu, and G.~Martius, ``"control what you can:
  Intrinsically motivated hierarchical reinforcement learner",'' in Deep RL
  Workshop NeurIPS, 2018.

\bibitem{narvekar2018learning}
S.~Narvekar and P.~Stone, ``Learning curriculum policies for reinforcement
  learning,'' \emph{arXiv preprint arXiv:1812.00285}, 2018.

\bibitem{weinshall2018curriculum}
D.~Weinshall and G.~Cohen, ``Curriculum learning by transfer learning: Theory
  and experiments with deep networks,'' \emph{arXiv preprint arXiv:1802.03796},
  2018.

\bibitem{stout2010competence}
A.~Stout and A.~G. Barto, ``Competence progress intrinsic motivation,'' in
  \emph{2010 IEEE 9th International Conference on Development and
  Learning}.\hskip 1em plus 0.5em minus 0.4em\relax IEEE, 2010, pp. 257--262.

\bibitem{baranes2010intrinsically}
A.~Baranes and P.-Y. Oudeyer, ``\BIBforeignlanguage{English}{Intrinsically
  motivated goal exploration for active motor learning in robots: A case
  study},'' in \emph{\BIBforeignlanguage{English}{{IEEE/RSJ International
  Conference on Intelligent Robots and Systems (IROS 2010)}}}.\hskip 1em plus
  0.5em minus 0.4em\relax Taipei, Taiwan, Province Of China: IEEE, 2010.

\bibitem{veeriah2018many}
V.~Veeriah, J.~Oh, and S.~Singh, ``Many-goals reinforcement learning,''
  \emph{arXiv preprint arXiv:1806.09605}, 2018.

\bibitem{colas2018curious}
C.~Colas, P.~Fournier, O.~Sigaud, and P.-Y. Oudeyer, ``{CURIOUS}: Intrinsically
  motivated multi-task, multi-goal reinforcement learning,'' \emph{arXiv
  preprint arXiv:1810.06284}, 2018.

\bibitem{nguyen2011bootstrapping}
S.~M. Nguyen, A.~Baranes, and P.-Y. Oudeyer, ``Bootstrapping intrinsically
  motivated learning with human demonstrations,'' \emph{arXiv preprint
  arXiv:1112.1937}, 2011.

\bibitem{thomas2017independently}
V.~Thomas, J.~Pondard, E.~Bengio, M.~Sarfati, P.~Beaudoin, M.-J. Meurs,
  J.~Pineau, D.~Precup, and Y.~Bengio, ``Independently controllable features,''
  \emph{arXiv preprint arXiv:1708.01289}, 2017.

\bibitem{machado2017revisiting}
M.~C. Machado, M.~G. Bellemare, E.~Talvitie, J.~Veness, M.~Hausknecht, and
  M.~Bowling, ``Revisiting the arcade learning environment: Evaluation
  protocols and open problems for general agents,'' \emph{arXiv preprint
  arXiv:1709.06009}, 2017.

\bibitem{van2015deep}
H.~Van~Hasselt, A.~Guez, and D.~Silver, ``Deep reinforcement learning with
  double {Q}-learning,'' \emph{CoRR, abs/1509.06461}, 2015.

\bibitem{baranes2013active}
A.~Baranes and P.-Y. Oudeyer, ``Active learning of inverse models with
  intrinsically motivated goal exploration in robots,'' \emph{Robotics and
  Autonomous Systems}, vol.~61, no.~1, pp. 49--73, 2013.

\bibitem{kang2018policy}
B.~Kang, Z.~Jie, and J.~Feng, ``Policy optimization with demonstrations,'' in
  \emph{International Conference on Machine Learning}, 2018, pp. 2474--2483.

\bibitem{vevcerik2017leveraging}
M.~Ve{\v{c}}er{\'\i}k, T.~Hester, J.~Scholz, F.~Wang, O.~Pietquin, B.~Piot,
  N.~Heess, T.~Roth{\"o}rl, T.~Lampe, and M.~Riedmiller, ``Leveraging
  demonstrations for deep reinforcement learning on robotics problems with
  sparse rewards,'' \emph{arXiv preprint arXiv:1707.08817}, 2017.

\end{thebibliography}

\end{document}